\documentclass[runningheads]{llncs}
\usepackage{graphicx}
\usepackage{xcolor}
\usepackage{bm}
\usepackage{amsmath}
\usepackage{amsfonts}
\usepackage{float}
\usepackage{booktabs}
\usepackage{multirow} 
\usepackage{url}
\usepackage{hyperref}
\usepackage{cite}
\usepackage{array}
\usepackage{verbatim}
\usepackage{color}
\usepackage{marvosym}

\makeatletter
\newcommand{\printfnsymbol}[1]{%
  \textsuperscript{\@fnsymbol{#1}}%
}
\makeatother

\begin{document}
\title{Positive-unlabeled Learning for Cell Detection in Histopathology Images with Incomplete Annotations}
%\title{Cell Detection in Histopathology Images with Incomplete Annotations}
\author{Zipei Zhao$^{\mathrm{1},}$\thanks{Equal contribution; \textsuperscript{\Letter} co-corresponding authors.}, Fengqian Pang$^{2,}$\printfnsymbol{1}, Zhiwen Liu$^{1}$\textsuperscript{(\Letter)}, Chuyang Ye$^{1}$\textsuperscript{(\Letter)}}
% index{Zhao, Zipei}
% index{Pang, Fengqian}
% index{Liu, Zhiwen}
% index{Ye, Chuyang}

\authorrunning{Zhao and Pang et al.}
\titlerunning{PU Learning for Cell Detection}
\institute{$^{1}$School of Information and Electronics, Beijing Institute of Technology, Beijing, China\\
\email{\{zwliu,chuyang.ye\}@bit.edu.cn}\\
$^{2}$School of Information Science and Technology, North China University of Technology, Beijing, China}
\maketitle              % typeset the header of the contribution
\begin{abstract}
Cell detection in histopathology images is of great value in clinical practice. \textit{Convolutional neural networks} (CNNs) have been applied to cell detection to improve the detection accuracy, where cell annotations are required for network training. However, due to the variety and large number of cells, complete annotations that include every cell of interest in the training images can be challenging. Usually, incomplete annotations can be achieved, where positive labeling results are carefully examined to ensure their reliability but there can be other positive instances, i.e., cells of interest, that are not included in the annotations. This annotation strategy leads to a lack of knowledge about true negative samples.
Most existing methods simply treat instances that are not labeled as positive as truly negative during network training, which can adversely affect the network performance. 
In this work, to address the problem of incomplete annotations, we formulate the training of detection networks as a positive-unlabeled learning problem. 
Specifically, the classification loss in network training is revised to take into account incomplete annotations, where the terms corresponding to negative samples are approximated with the true positive samples and the other samples of which the labels are unknown. 
To evaluate the proposed method, experiments were performed on a publicly available dataset for mitosis detection in breast cancer cells, and the experimental results show that our method improves the performance of cell detection given incomplete annotations for training.

\keywords{Cell detection \and positive-unlabeled learning \and incomplete annotation}
\end{abstract}
\section{Introduction}

Clinical medicine relies largely on cell detection and counting in histopathology images to assess the degree of tissue damage. 
However, conventional detection methods~\cite{veta2014breast} are complicated and inefficient, and they cannot cope with the large amount of image data. In recent years, the application of \textit{convolutional neural networks}~(CNNs) to object detection has shown unparalleled advantages, and CNN-based algorithms are increasingly used in the cell detection task with remarkable results~\cite{mitosis,zongshu1,zongshu5,zongshu3}. 

For training CNN-based detectors, high-quality annotations (bounding boxes in most cases) of the cells of interest are needed. Ideally, every cell of interest in the training images should be annotated by experts. 
However, due to the large number and diverse morphology of cells in histopathology images, completely annotating all instances becomes very challenging. In practice, experts can prefer to only ensure that all instances labeled as positive are correct~\cite{dataset}, and the annotation may even be sparse in the image if a large number of images are to be annotated~\cite{bde}. In this case, the rest of the instances may not be all true negatives, and there still exist unannotated positive instances with high probability. In other words, the annotations are incomplete and only include a subset of the cells of interest.  Most existing methods simply treat unannotated areas as negative instances during network training \cite{isbiii}. 
Although this training strategy may still achieve promising results given a relatively large number of annotated positive instances, it neglects the existence of false negative training data due to incomplete annotations and is thus suboptimal. 
%Although they have achieved promising results due to the relatively large number of annotated positive instances, such a training strategy neglects the false negative training data due to incomplete annotations and is thus suboptimal. 

Very recently, the problem of incomplete annotations for training CNN-based cell detectors draws attention~\cite{bde}. In \cite{bde}, motivated by the significant density difference between the regression boxes of positive and negative instances, \textit{Boxes Density Energy}~(BDE) is proposed to calibrate the training loss. 
The assumption of BDE is that negative samples are inclined to possess a lower density of regression boxes. Therefore, among the instances that are not annotated as positive, the regions with a higher density of regression boxes are less likely to be truly negative, and their losses as negative samples are calibrated to be smaller.
Compared with methods that neglect the problem of incomplete annotations, the detection performance is improved with BDE. 
To the best of our knowledge, this is the only existing work that addresses the problem of incomplete annotations for cell detection\footnote{The work in~\cite{dataset} requires the annotated mask of each instance in addition to the bounding box, and thus it addresses a different problem.}, and the development of cell detection methods with incomplete annotations is still an open problem.

In this work, we continue to explore the problem of incomplete annotations for CNN-based cell detection in histopathology images. Since the regions without annotated instances may include both positive and negative samples, we treat the instances in these regions as unlabeled data and propose to reformulate the classification part of the detection problem as a \textit{positive-unlabeled} (PU) learning problem~\cite{pu1,pu2}.
In particular, the classification loss for network training is revised, where the terms about negative samples are approximated with the annotated positive instances and unlabeled instances. 
The revised classification loss is then integrated with the localization loss to train the detection network.
%In particular, the classification loss for network training in typical classification problems is revised, where the terms about negative samples are approximated with the annotated positive instances and unlabeled instances. 
%Therefore, we further incorporate negative training samples by 1) exploiting images without cells of interest, e.g., images of normal tissue for tumor cell detection, and 2) more thoroughly exploring the unlabeled samples in the incomplete annotations using the BDE technique that is complementary to the proposed PU learning framework.
%To evaluate the proposed method, we performed experiments on the publicly available MITOS-ATYPIA-14 dataset\footnote{\url{https://mitos-atypia-14.grand-challenge.org/}}, which targets the detection of mitosis in breast cancer cells.
To evaluate the proposed method, we performed experiments on a publicly available dataset for mitosis detection in breast cancer histopathology images.
For demonstration, Faster R-CNN~\cite{fasterrcnn} was used as our backbone detection network, which has been previously applied to various cell detection tasks~\cite{isbiii,celldete}. Experimental results show that our method leads to improved cell detection performance given incomplete annotations for training.

\section{Methods}

\subsection{Background: cell detection with complete annotations for training}

In typical deep learning based methods of object detection, e.g., Faster R-CNN~\cite{fasterrcnn}, the network generates a bounding box to indicate the position of each instance and the corresponding class probability, i.e., the likelihood of the instance belonging to a certain category. 
In this work, we are interested in cell detection, which usually aims at the detection of a certain type of cells, e.g., cells related to cancer~\cite{mitosis,zongshu1,tucell}. Thus, we assume that the classification is binary, i.e., the ground truth label $z$ of the bounding box $x$ is binary: $z\in\{0,1\}$.
Note that $x$ is generally initially produced by a region proposal module, such as a region proposal network~\cite{fasterrcnn}.
The localization of the bounding box $x$ given by the network is denoted by $v=\{X,Y,W,H\}$, where $X$, $Y$, $W$, and $H$ represent the $x$-coordinate, $y$-coordinate, width, and height of the bounding box, respectively. The probability of a bounding box being positive---i.e., $z=1$---predicted by the network is denoted by $c\in[0,1]$.

Conventionally, to train a detection network, all positive instances should be annotated for the training images, and the loss comprising both localization and classification error is minimized. 
The localization loss $\mathcal{L}_{\mathrm{loc}}$ is computed from the predicted location $v$ and the ground truth location $b=\{X_{b},Y_{b},W_{b},H_{b}\}$ of the positive instances. For example, a typical choice of $\mathcal{L}_{\mathrm{loc}}$ is the smooth $L_{1}$ loss function~\cite{bde}, where
\begin{eqnarray}
\mathcal{L}_{\mathrm{loc}}=\sum\limits_{i=1}^{N_{\mathrm{p}}}\sum\limits_{\substack{(v^{i},b^{i})\in  \{(X^{i},X_{b}^{i}),(Y^{i},Y_{b}^{i}), \\ (W^{i},W_{b}^{i}),(H^{i},H_{b}^{i})\}}}\mathrm{smooth}_{L_{1}}(v^{i}-b^{i})\\
\mathrm{with}\quad 
\mathrm{smooth}_{L_{1}}(a)=\left\{ {\begin{array}{*{20}{c}}
		{{{a}^2}/2,} & {\mathrm{if}\left| {a} \right| \le 1}\\
		{\left| {a} \right| - 0.5,}&{\mathrm{otherwise}}
\end{array}} \right ..
\label{eq:regression}
\end{eqnarray}
Here, $i$ and $N_{\mathrm{p}}$ are the index and the total number of positive training samples, respectively.
The classification loss $\mathcal{L}_{\mathrm{cls}}$ is computed from the predicted classification probability and the corresponding ground truth label as follows
\begin{eqnarray}
\mathcal{L}_{\mathrm{cls}}=\frac{1}{N_{\mathrm{n}}+N_{\mathrm{p}}}\left(\sum\limits_{j=1}^{N_{\mathrm{n}}} H(c^{j}_{\mathrm{n}},0)  + \sum\limits_{i=1}^{N_{\mathrm{p}}} H(c^{i}_{\mathrm{p}},1)\right).
\label{eq:cls}
\end{eqnarray}
Here, $j$ and $N_\mathrm{n}$ are the index and the total number of negative training samples (samples that have no overlap or do not have a sufficiently large overlap with the labeled positive instances), respectively; $c^{i}_{\mathrm{p}}$ and $c^{j}_{\mathrm{n}}$ represent the predicted probability of the positive sample $x^{i}_{\mathrm{p}}$ and the negative sample $x^{j}_{\mathrm{n}}$ being positive, respectively; and most commonly $H(\cdot,\cdot)$ is a cross-entropy loss function that measures the difference between the prediction and ground truth~\cite{cross}. 
With the complete annotations where every positive instance in the training images is labeled, the sum of the two losses $\mathcal{L}_{\mathrm{loc}}$ and $\mathcal{L}_{\mathrm{cls}}$ is minimized to learn the weights of the detection network.

\subsection{PU learning for cell detection with incomplete annotations}
\label{sec:pu}

For cell annotations on histopathology images, because there are usually a huge number of cells with various appearances, it is challenging to annotate every positive instance. 
Experts may only ensure that the annotated cells are truly positive, and the annotated cells may even appear sparse in the image to reduce the annotation load~\cite{bde}.
In this case, the training set only contains a subset of the positive instances and misses other positive instances. 
In other words, in this incompletely annotated dataset, the regions with no instances labeled as positive are not necessarily all truly negative. Therefore, given incomplete annotations, training the detection network with the classification loss described in Eq.~(\ref{eq:cls}) is no longer accurate and could cause performance degradation. 

Since the regions that are not labeled as positive may comprise both positive and negative samples, the instances in these regions can be considered unlabeled. This means that the incompletely annotated training dataset contains both positively labeled and unlabeled training samples ($x_{\mathrm{p}}$ and $x_{\mathrm{u}}$, respectively). Thus, to address the problem of incomplete annotations for cell detection, we propose to exploit PU learning, so that the classification loss that is originally computed with complete annotations can be approximated with incomplete annotations.

To derive the approximation of the classification loss, we notice that $\mathcal{L}_{\mathrm{cls}}$ is an approximation (empirical mean) of the expectation $\mathbb{E}_{(x,z)}[H(c,z)]$ that measures the difference between the predicted probability $c$ of $x$ being a positive sample and the ground truth label $z$. The computation of $\mathbb{E}_{(x,z)}[H(c,z)]$ can be reformulated as
\begin{eqnarray}
&& \mathbb{E}_{(x,z)}[H(c,z)] \nonumber\\
&=&\mathrm{Pr}(z=0) \int p(x|z=0)H(c,0)\mathrm{d}x + \mathrm{Pr}(z=1)\int p(x|z=1)H(c,1)\mathrm{d}x \\ 
&=&(1-\pi)\mathbb{E}_{x|z=0}[H(c,0)] +\pi \mathbb{E}_{x|z=1}[H(c,1)].
\label{eq:reform}
\end{eqnarray}
Here, $p(\cdot)$ represents a probability density function, and $\pi=\mathrm{Pr}(z=1)$ is the positive class prior. 

Since positive training samples are available yet negative training samples are unavailable, we can directly compute the second term in Eq.~(\ref{eq:reform}) with the training samples but not the first term. However, it is possible to approximate the first term with both positive and unlabeled training samples~\cite{nopu}. Because $p(x)=\mathrm{Pr}(z=0)p(x|z=0) + \mathrm{Pr}(z=1)p(x|z=1)$, we have 
\begin{eqnarray}
\mathrm{Pr}(z=0)p(x|z=0) = p(x) - \mathrm{Pr}(z=1)p(x|z=1),
\end{eqnarray}
and the first term $(1-\pi)\mathbb{E}_{x|z=0}[H(c,0)]$ in Eq.~(\ref{eq:reform}) becomes
\begin{eqnarray}
&&\mathrm{Pr}(z=0) \int p(x|z=0)H(c,0)\mathrm{d}x \nonumber\\
&=&\int p(x) H(c,0)\mathrm{d}x -\mathrm{Pr}(z=1) \int p(x|z=1) H(c,0)\mathrm{d}x \\
&=& \mathbb{E}_{x}[H(c,0)] - \pi \mathbb{E}_{x|z=1}[H(c,0)].
\label{eq:approx}
\end{eqnarray}
Then, based on Eq.~(\ref{eq:reform}), $\mathbb{E}_{(x,z)}[H(c,z)]$ becomes
\begin{eqnarray}
\mathbb{E}_{(x,z)}[H(c,z)] = \mathbb{E}_{x}[H(c,0)] - \pi \mathbb{E}_{x|z=1}[H(c,0)] +\pi \mathbb{E}_{x|z=1}[H(c,1)].
\label{eq:pu_exp}
\end{eqnarray}
Now, the second and third terms on the right hand side of Eq.~(\ref{eq:pu_exp}) can be computed with positive samples, and the first term $\mathbb{E}_{x}[H(c,0)]$ still needs to be determined.

In PU learning for classification problems, it is assumed that the distribution of the unlabeled data $x_{\mathrm{u}}$ is the same as the distribution of $x$, so that $\mathbb{E}_{x}[H(c,0)]$ can be approximated by $\mathbb{E}_{x_{\mathrm{u}}}[H(c,0)]$. Previous work has directly applied such approximation to an object detection problem~\cite{Yang2020object}. However, simply applying the PU learning strategy developed for classification to the detection problem can be problematic, because in detection problems the unlabeled samples and positively labeled samples originate from the same images. 
Some positive samples are excluded from the distribution of $x_{\mathrm{u}}$, and thus the approximation is biased. 
Instead, if we combine the positively labeled and unlabeled samples in the same images, they provide samples drawn from the distribution of $x$, and $\mathbb{E}_{x}[H(c,0)]$ can be approximated using the combination of these samples:
\begin{eqnarray}
\mathbb{E}_{x}[H(c,0)] \approx \frac{1}{N_{\mathrm{u}} + N_{\mathrm{p}}}\left(\sum_{k=1}^{N_{\mathrm{u}}}H(c_{\mathrm{u}}^{k},0) + \sum_{i=1}^{N_{\mathrm{p}}}H(c_{\mathrm{p}}^{i},0) \right),
\label{eq:approx_unlabeled}
\end{eqnarray}
where $k$ and $N_{\mathrm{u}}$ are the index and the total number of unlabeled training samples, respectively, and $c^{k}_{\mathrm{u}}$ represents the predicted probability of the unlabeled sample~$x^{k}_{\mathrm{u}}$.
In this way, all terms on the right hand side of Eq.~(\ref{eq:pu_exp}) can be approximated with the training samples.
Note that in practice an expressive CNN may overfit the data and produce negative values for the approximation of $(1-\pi)\mathbb{E}_{x|z=0}[H(c,0)]$ in Eq.~(\ref{eq:approx}). 
Thus, we follow \cite{nopu} and use a nonnegative approximation of $(1-\pi)\mathbb{E}_{x|z=0}[H(c,0)]$, which leads to
\begin{eqnarray}
&&\mathbb{E}_{x}[H(c,0)] - \pi \mathbb{E}_{x|z=1}[H(c,0)] \nonumber\\
&\approx& \max \left\{0, \frac{1}{N_{\mathrm{u}} + N_{\mathrm{p}}}\left(\sum_{k=1}^{N_{\mathrm{u}}}H(c_{\mathrm{u}}^{k},0) + \sum_{i=1}^{N_{\mathrm{p}}}H(c_{\mathrm{p}}^{i},0) \right)  - \frac{\pi}{N_{\mathrm{p}}}\sum_{i=1}^{N_{\mathrm{p}}}H(c_{\mathrm{p}}^{i},0)\right\} .\nonumber\\
&&\mbox{ }
\end{eqnarray}
Then, we have the revised classification loss that approximates $\mathbb{E}_{(x,z)}[H(c,z)]$ with the PU learning framework:
\begin{eqnarray}
\mathcal{L}_{\mathrm{cls}}^{\mathrm{pu}} &=& \max \left\{0, \frac{1}{N_{\mathrm{u}} + N_{\mathrm{p}}}\left(\sum_{k=1}^{N_{\mathrm{u}}}H(c_{\mathrm{u}}^{k},0) + \sum_{i=1}^{N_{\mathrm{p}}}H(c_{\mathrm{p}}^{i},0) \right)  - \frac{\pi}{N_{\mathrm{p}}}\sum_{i=1}^{N_{\mathrm{p}}}H(c_{\mathrm{p}}^{i},0)\right\} \nonumber\\
&&+ \frac{\pi}{N_{\mathrm{p}}}\sum_{i=1}^{N_{\mathrm{p}}}H(c^{i}_{{\mathrm{p}}},1).
\end{eqnarray}
Note that the class prior $\pi$ in $\mathcal{L}_{\mathrm{cls}}^{\mathrm{pu}}$ is assumed to be known. Since it is difficult to directly estimate $\pi$ using incompletely annotated training samples, $\pi$ can be considered a hyperparameter and determined with a validation set (see Sect.~\ref{sec:acc}).

With the revised classification loss, the complete loss function for training the detection network with incomplete annotations is
\begin{eqnarray}
\mathcal{L} = \mathcal{L}_{\mathrm{loc}} + \mathcal{L}_{\mathrm{cls}}^{\mathrm{pu}}.
\label{eq:all}
\end{eqnarray}
This loss function can be integrated with different state-of-the-art backbone detection networks that are based on the combination of localization and classification losses, e.g., Faster R-CNN~\cite{fasterrcnn} that is widely applied to object detection problems including cell detection.

\section{Results}
\label{sec:results}
\subsection{Dataset description and experimental settings}

To evaluate the proposed method, we performed experiments on the publicly available MITOS-ATYPIA-14 dataset\footnote{\url{https://mitos-atypia-14.grand-challenge.org/}}, which aims to detect mitosis in breast cancer cells. 
The dataset comprises 393 images, and the image size is 1600$\times $1600. 
In this dataset, experienced pathologists have annotated each mitosis with a key point, and like~\cite{bde} for each key point we generated a 32$\times $32 bounding box centered around it. 

The images were split into a training, validation, and test set with a ratio of about 4:1:1, and we performed 5-fold cross-validation for evaluation, where the validation set was fixed and the training and test sets were regrouped in each fold. 
Since the size of the original image is large, we cropped the images into $500\times 500$ patches with an overlap of 100 pixels horizontally and vertically between adjacent patches.
In addition, to simulate the scenario where incomplete annotations are performed, like~\cite{bde} for each image in the training or validation set, we randomly deleted the annotations until there was only one annotation per image patch. 
The annotations in the test set were intact.

For demonstration, we used Faster R-CNN \cite{fasterrcnn} as the baseline network to detect the mitosis, as it has been previously used for similar detection tasks~\cite{isbiii,celldete}. The Faster R-CNN had been pretrained on ImageNet~\cite{imagenet} for a better initialization of network weights.\footnote{The VGG16~\cite{vgg} backbone was selected here, but similar results (not reported) were achieved with the ResNet50 and ResNet101~\cite{Resnet} backbones.}
For each patch, at most 100 prediction boxes were generated according to the confidence score (greater than 0.5) in descending order~\cite{fasterrcnn}. 
For test images, the detection result on each patch was merged to produce the final prediction. 
Specifically, we generated prediction boxes for each $500\times500$ patch, mapped the coordinates of these boxes back into the image, and performed non-maximum suppression~\cite{nms} to merge duplicate bounding boxes.

The Adam optimizer~\cite{adam} was used for minimizing the loss function, where the initial learning rate was set to $10^{-3}$. To ensure training convergence, the detection network was trained with 20 epochs, and the training procedure took about 4 hours on an NVIDIA GeForce GTX 1080 Ti GPU. 

\subsection{Detection accuracy}
\label{sec:acc}

As described in Sect.~\ref{sec:pu}, the class prior $\pi$ was determined based on the validation set. 
The candidate values of $\pi$ ranged from 0.02 to 0.06 with an increment of 0.01.
Because there were only incompletely annotated images in the validation set, precision could not be used to evaluate the performance on the validation set~\cite{isbiii}. Thus, we selected $\pi$ according to the best average recall computed from the validation set. Note that $\pi$ was selected for each fold independently, and the selected value was consistent (0.04 or 0.05) across the folds.

We compared the proposed method with two competing methods, which, for fair comparison, used the same backbone Faster R-CNN detection network. 
In the first competing method, the Faster R-CNN model (initialized with the pretraining on ImageNet) was trained using the incomplete annotations, where the unlabeled regions were simply considered negative. This Faster R-CNN is referred to as the baseline method. In the second competing method, we integrated the BDE method~\cite{bde} with the baseline Faster R-CNN, so that weights were added to the unlabeled samples to calibrate the classification loss according to the density of regression boxes.
Note that in the BDE method, the samples in the unlabeled areas are still considered negative, but their weights are adjusted based on how likely they are to be really negative, whereas in our method the samples in the unlabeled areas are considered to have unknown classes, and they are used together with positively labeled samples to approximate the classification loss in a principled framework.  

\begin{figure}[!t]
	\centering
	\includegraphics[width=0.85\textwidth]{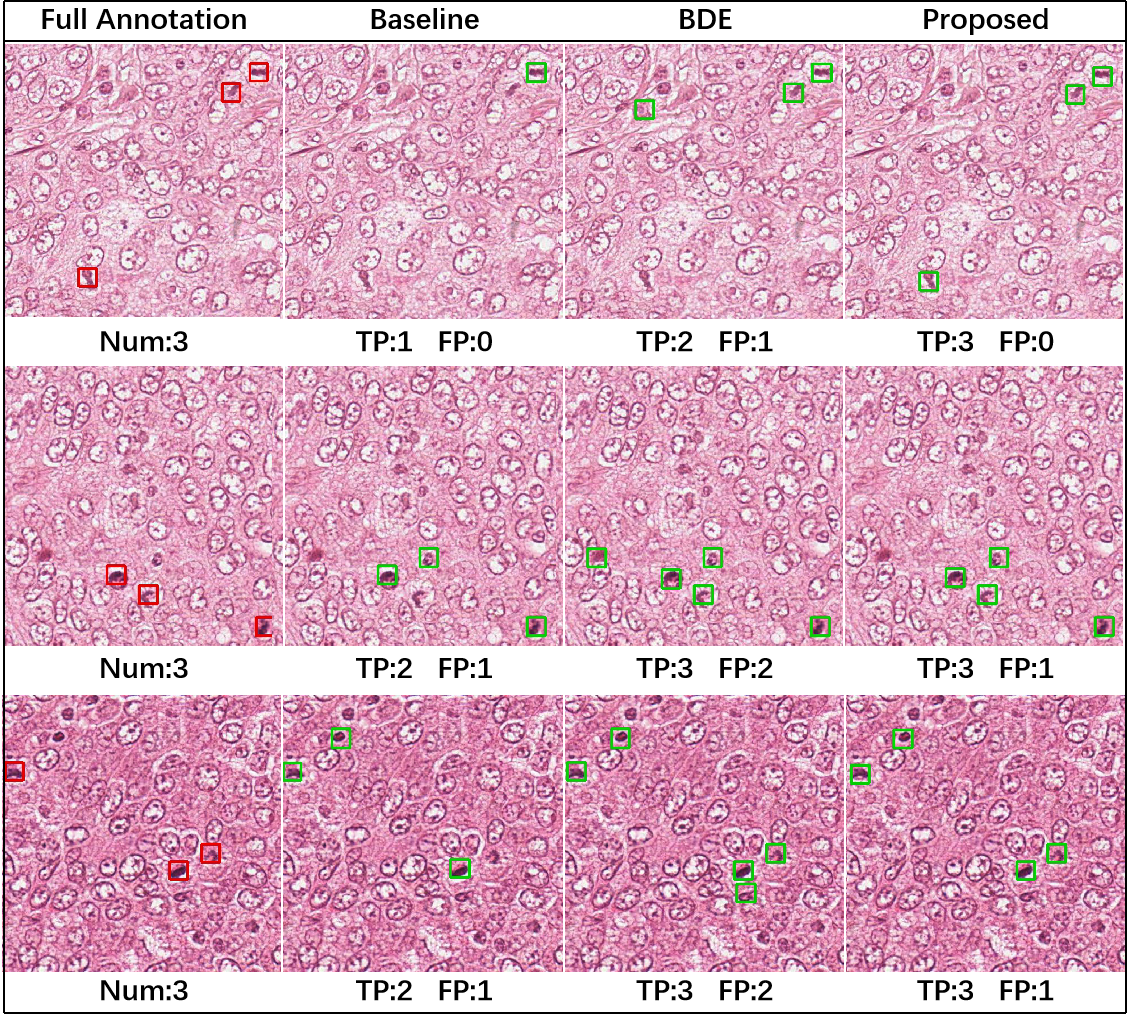}
	\caption{Examples of detection results on test patches shown together with the full annotations. The numbers of TP and FP detection results and the number of mitoses in the annotation are also indicated for each case.} \label{fig1}
\end{figure}

We first qualitatively evaluated the proposed method.
Examples of the detection results of each method on test patches are shown in Fig.~\ref{fig1}, together with the full annotations. 
The numbers of \textit{true positive}~(TP) and \textit{false positive}~(FP) detection results are also indicated in the figure for each case.
In these cases, our method compares favorably with the competing methods by either producing more TP boxes than the competing methods without increasing the number of FP boxes or reducing the number of FP boxes with preserved TP boxes.

\begin{table}[!t]
	\centering
	\caption{The average recall and average precision on the test set for each fold. The results of the proposed method are highlighted in bold.}
	\label{tab1}
	\resizebox{\textwidth}{10mm}{
		\begin{tabular}{c >{\centering\arraybackslash}p{1.2cm} >{\centering\arraybackslash}p{1.2cm} >{\centering\arraybackslash}p{1.2cm} >{\centering\arraybackslash}p{1.2cm} >{\centering\arraybackslash}p{1.2cm} >{\centering\arraybackslash}p{1.2cm} >{\centering\arraybackslash}p{1.2cm} >{\centering\arraybackslash}p{1.2cm} >{\centering\arraybackslash}p{1.2cm} >{\centering\arraybackslash}p{1.2cm}}
			\hline
	  	    \multirow{2}{*}{Method} & \multicolumn{2}{c }{Fold 1}& \multicolumn{2}{c }{Fold 2}&\multicolumn{2}{c }{Fold 3}&\multicolumn{2}{c }{Fold 4}& \multicolumn{2}{c}{Fold 5} \\
		  	\cline{2-11} 
			         			    &    Recall & Precision        &      Recall &  Precision     &   Recall  & Precision       &     Recall  &  Precision    &   Recall  &  Precision     \\
			\hline
			\hline
			Baseline 				&    0.778  		&  0.533          	&      0.733  			&   0.506       		&    0.753  &   0.543        &      0.710  &   0.510      &    0.649  &  0.412   \cr
			BDE      					&    0.821  		&  0.552          	&      0.792  			&   0.529       		&    0.789  &   0.557        &      0.725  &    0.528     &    0.670  &  0.415   \cr
			Proposed 				&    \bf{0.842}	&  \bf{0.567}		&      \bf{0.802}  	&   \bf{0.547}    &    \bf{0.801}  &   \bf{0.559}  & \bf{0.738}  & \bf{0.541}  & \bf{0.670} & \bf{0.435}    \\
			\hline
	\end{tabular}}
\end{table}

Next, we quantitatively compared the proposed method with the competing methods.
We computed the average recall and average precision of the detection results on the test set for each fold, and they are shown in Table~\ref{tab1}. 
Compared with the competing methods, the proposed method has higher recall and precision values, which indicate the better detection accuracy of the proposed method.

We also computed the means and standard deviations of the average recall and average precision of the five folds, and compared the proposed method with the competing methods using paired Student's $t$-tests. These results are shown in Table~\ref{tab:tab2}. Consistent with Table~\ref{tab1}, the proposed method has higher recall and precision. In addition, the improvement of our method is statistically significant.

\begin{table}[!t]
	\centering
	\caption{The means and \textit{standard deviations}~(stds) of the average recall and average precision of the five folds. The results of the proposed method are highlighted in bold. Asterisks indicate that the difference between the proposed method and the competing method is significant using a paired Student's $t$-test. ($^{*}p<0.05$, $^{**}p<0.01$)}
	\label{tab:tab2}
	\setlength{\tabcolsep}{1.5mm}{
	\scalebox{0.92}{
		\begin{tabular}{c >{\centering\arraybackslash}p{1.2cm} >{\centering\arraybackslash}p{1cm} >{\centering\arraybackslash}p{1.4cm} >{\centering\arraybackslash}p{1.2cm} >{\centering\arraybackslash}p{1cm} >{\centering\arraybackslash}p{1.4cm} >{\centering\arraybackslash}p{1.2cm} >{\centering\arraybackslash}p{1.2cm} >{\centering\arraybackslash}p{1.2cm} >{\centering\arraybackslash}p{1.2cm}}
			\hline
			\multirow{1}{*}{} & \multicolumn{3}{c }{Recall}& \multicolumn{3}{c }{Precision} \\
			\cmidrule(r){2-4} \cmidrule(r){5-7}
			& Baseline &  BDE &  Proposed  & Baseline  & BDE  & Proposed        \\
			\hline
			\hline
			mean & 0.725 & 0.759   & \bf{0.771}   & 0.501   & 0.516 & \bf{0.530}  \\ 
			%			\hline 
			std & 0.044  & 0.055 & \bf{0.060}  & 0.047  & 0.052 & \bf{0.048} \\
			%			\hline
			$p$ & ** & *  & - & **  & * & -\\
			\hline
	\end{tabular}}}
\end{table}

Finally, to confirm the benefit of the approximation developed in Eq.~(\ref{eq:approx_unlabeled}) for detection problems, we performed experiments with the original PU learning strategy for classification problems as in~\cite{Yang2020object} for comparison. The means of the average recall and average precision of the five folds were computed, which are 0.770 and 0.503, respectively.
Although the recall is comparable to the result (0.771) of the proposed method, the mean precision is much worse than the result (0.530) of the proposed method (see Table~\ref{tab:tab2}). The mean precision is even worse than the BDE result (0.516) and close to the baseline result (0.501) reported in Table~\ref{tab:tab2}. These comparisons confirm that directly applying the PU learning strategy developed for classification problems may not be suitable for cell detection in histopathology images.

\section{Conclusion}

In this work, we seek to address the problem of network training with incomplete annotations for cell detection in histopathology images. We propose to apply PU learning to cell detection, so that the classification loss is more appropriately computed from the incompletely annotated data during network training. The experimental results on a publicly available dataset show that our method can improve the performance of cell detection given incomplete annotations.

\subsubsection{Acknowledgements}

This work is supported by National Natural Science Foundation of China (62001009).

\bibliographystyle{splncs04}
\bibliography{refs}

\begin{thebibliography}{10}
\providecommand{\url}[1]{\texttt{#1}}
\providecommand{\urlprefix}{URL }
\providecommand{\doi}[1]{https://doi.org/#1}

\bibitem{pu2}
Charles, E., Keith, N.: Learning classifiers from only positive and unlabeled
  data. In: International Conference on Knowledge Discovery and Data Mining.
  pp. 213--220 (2008)

\bibitem{mitosis}
Dan, C., Alessandro, G., Bardella, L.: Mitosis detection in breast cancer
  histology images with deep neural networks. In: International Conference on
  Medical Image Computing and Computer-Assisted Intervention. pp. 411--418.
  Springer (2013)

\bibitem{imagenet}
Deng, J., Dong, W., Socher, R., Li, L.J., Li, K., Fei-Fei, L.: Image{Net}: A
  large-scale hierarchical image database. In: IEEE Conference on Computer
  Vision and Pattern Recognition. pp. 248--255 (2009)

\bibitem{pu1}
Fabien, L., Francois, D., Remi, G.: Learning from positive and unlabeled
  examples. In: Algorithmic Learning Theory. pp. 71--85 (2000)

\bibitem{Resnet}
He, K., Zhang, X., Ren, S., Sun, J.: Deep residual learning for image
  recognition. In: IEEE Conference on Computer Vision and Pattern Recognition.
  pp. 770--778 (2016)

\bibitem{vgg}
Karen, S., Andrew, Z.: Very deep convolutional networks for large-scale image
  recognition. arXiv preprint arXiv:1409.1556  (2015)

\bibitem{adam}
Kingma, D.P., Ba, J.: Adam: A method for stochastic optimization. arXiv
  preprint arXiv:1412.6980  (2014)

\bibitem{nopu}
Kiryo, R., Niu, G., Marthinus, P., Masashi, S.: Positive-unlabeled learning
  with non-negative risk estimator. In: Advances In Neural Information
  Processing Systems. pp. 1674--1684 (2017)

\bibitem{bde}
Li, H., Han, X., Kang, Y., Shi, X., Yan, M., Tong, Z., Bu, Q., Cui, L., Feng,
  J., Yang, L.: A novel loss calibration strategy for object detection networks
  training on sparsely annotated pathological datasets. In: International
  Conference on Medical Image Computing and Computer-Assisted Intervention. pp.
  320--329. Springer (2020)

\bibitem{dataset}
Li, J., Yang, S., Huang, X., Da, Q., Yang, X., Hu, Z., Duan, Q., Wang, C., Li,
  H.: Signet ring cell detection with a semi-supervised learning framework. In:
  International Conference on Information Processing in Medical Imaging. pp.
  842--854 (2019)

\bibitem{nms}
Neubeck, A., Van~Gool, L.J.: Efficient non-maximum suppression. In:
  International Conference on Pattern Recognition. pp. 850--855 (2006)

\bibitem{cross}
Phan, T.H., Yamamoto, K.: Resolving class imbalance in object detection with
  weighted cross entropy losses. arXiv preprint arXiv:2006.01413  (2020)

\bibitem{fasterrcnn}
Ren, S., He, K., Girshick, R., Sun, J.: Faster {R-CNN}: Towards real-time
  object detection with region proposal networks. In: Advances in Neural
  Information Processing Systems. pp. 91--99 (2015)

\bibitem{zongshu5}
Schmidt, U., Weigert, M., Broaddus, C., Myers, G.: Cell detection with
  star-convexpolygons. In: International Conference on Medical Image Computing
  and Computer-Assisted Intervention. pp. 265--273. Springer (2018)

\bibitem{tucell}
Sirinukunwattana, K., Raza, S.E.A., Tsang, Y.W., Snead, D.R., Cree, I.A.,
  Rajpoot, N.M.: Locality sensitive deep learning for detection and
  classification of nuclei in routine colon cancer histology images. IEEE
  Transactions on Medical Imaging  \textbf{35}(5),  1196--1206 (2016)

\bibitem{isbiii}
Sun, Y., Huang, X., Molina, E.G.L., Dong, L., Zhang, Q.: Signet ring cells
  detection in histology images with similarity learning. In: International
  Symposium on Biomedical Imaging. pp. 37--48 (2020)

\bibitem{zongshu3}
Tzu-Hsi, S., Victor, S., Hesham, E.: Simultaneous cell detection and
  classification in bone marrow histology images. IEEE Journal of Biomedical
  and Health Informatics  \textbf{23}(4),  1469--1476 (2018)

\bibitem{veta2014breast}
Veta, M., Pluim, J.P., Van~Diest, P.J., Viergever, M.A.: Breast cancer
  histopathology image analysis: A review. IEEE Transactions on Biomedical
  Engineering  \textbf{61}(5),  1400--1411 (2014)

\bibitem{zongshu1}
Xu, J., Lei, X., Liu, Q., Hannah, G., Wu, J.: Stacked sparse autoencoder
  ({SSAE}) for nuclei detection on breast cancer histopathology images. IEEE
  Transactions on Medical Imaging  \textbf{35}(1),  119--130 (2015)

\bibitem{Yang2020object}
Yang, Y., Liang, K.J., Carin, L.: Object detection as a positive-unlabeled
  problem. In: British Machine Vision Conference (2020)

\bibitem{celldete}
Zhang, J., Hu, H., Chen, S.: Cancer cells detection in phasecontrast microscopy
  images based on faster {R-CNN}. In: International Symposium on Computational
  Intelligence and Design. pp. 363--367 (2016)

\end{thebibliography}

\end{document}